\documentclass[]{aiaa-tc}
\pdfoutput=1

\usepackage{graphicx} 
\usepackage{hyperref}
\usepackage{amsmath}
\usepackage{siunitx}
\usepackage{subcaption}
\usepackage{algorithm2e}

 \title{$L_1$ guidance logic extension for small UAVs: handling high winds and small loiter radii}

 \author{
    Thomas Stastny \\
    \small \href{mailto:thomas.stastny@mavt.ethz.ch}{thomas.stastny@mavt.ethz.ch}\\%
    \small Autonomous Systems Lab, ETH Z\"{u}rich \\
    \small Z\"{u}rich, Switzerland \\
    \small Mar. 31, 2018 (Updated: May 22, 2018)
 }

 \AIAApapernumber{YEAR-NUMBER}
 \AIAAconference{Conference Name, Date, and Location}
 \AIAAcopyright{\AIAAcopyrightD{YEAR}}

\begin{document}

\maketitle

\section{Introduction}
$L_1$ guidance logic \cite{L1original} is one of the most widely used path following controllers for small fixed-wing unmanned aerial vehicles (UAVs), primarily due to its simplicity (low-cost implementation on embedded on-board processors, e.g. micro-controllers) and ability to track both circles and lines, which make up the vast majority of a typical fixed-wing vehicle's flight plan. 
In~\cite{L2plus}, the logic was extended to allow explicit setting of the $L_1$ period and damping, from which an adaptive the $L_1$ length and gain can be calculated, keeping dynamic similarity in the path convergence properties, independent of the velocity.
Two primary drawbacks remain, specific to small, slow flying fixed-wing UAVs:
\begin{enumerate}
\item As elaborated in~\cite{L1original}, circle following convergence requires that the $L_1$ length be less than or equal to the circle radius $R$, i.e. $L_1\leq R$. This condition may often be violated when the ground speed of the aircraft is high, or the circle radius is small.
\item $L_1$ logic breaks down when wind speeds exceed the vehicle's airspeed, another common predicament for small, slow-flying UAVs.
\end{enumerate}
Though many other guidance formulations exist which may partially address one or both of the listed issues, line-of-sight -based or otherwise, this brief limits its scope to presenting simple extensions to the extensively field tested $L_1$ guidance formulations commonly used in open source autopilot platforms today (e.g. \emph{PX4}\footnote{dev.px4.io} and \emph{Ardupiot}\footnote{\url{ardupilot.org}}), allowing legacy operators to keep existing controller tunings and still take advantage of the added performance and safety features.
First, an adaptive recalculation of the $L_1$ ratio is introduced in the event that the $L_1$ length violates the circle tracking convergence criteria.
Second, borrowing some concepts from~\cite{Furieri_ACC2017}, a bearing feasibility parameter is introduced which continuously transitions $L_1$ commands from the \emph{tracking} objective to a \emph{safety} objective, i.e. attempting to \emph{mitigate} run-away in over-wind scenarios.
Finally, an airspeed reference command increment is introduced which is non-zero according to the wind speed ratio (i.e. wind speed over airspeed) and the bearing feasibility, effectively \emph{preventing} run-away.
Focus will be kept to the circle following objective (or loiter), though the same principles may similarly be applied to waypoint or line tracking.
\section{Fundamentals}
\begin{figure}
  \centering
  \includegraphics[width=0.7\textwidth]{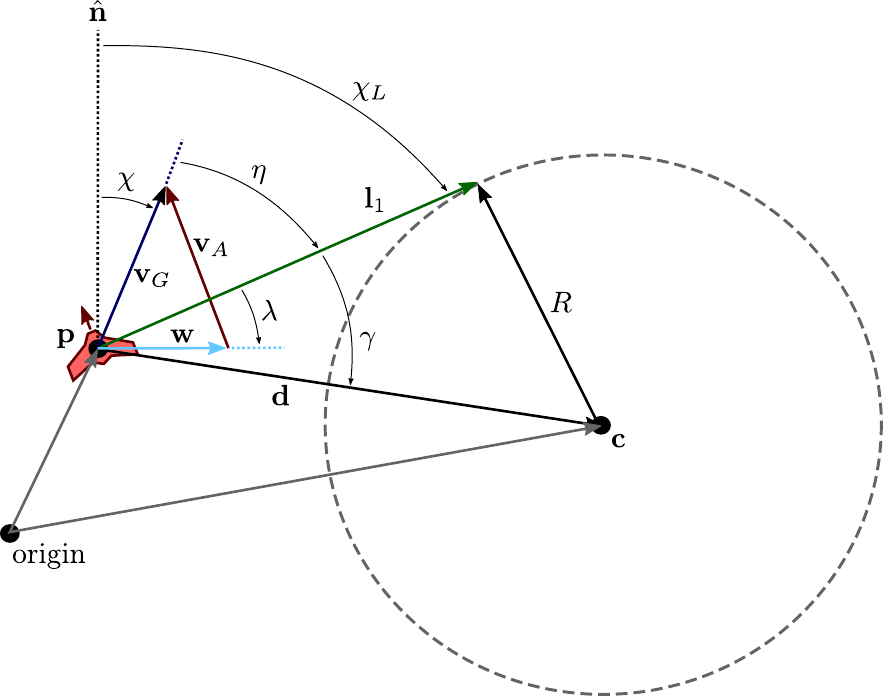}
  \caption{$L_1$ guidance geometry.}\label{fig:geom}
\end{figure}

A similar formulation to the $L_{2+}$ algorithm~\cite{L2plus} will form the basis of the extensions in this document.
Figure~\ref{fig:geom} shows the algorithm geometry and notation used throughout the following sections.
The original $L_1$ acceleration command, with ground speed as input, is formulated as:
\begin{equation}\label{eq:aref_orig}
a_\text{ref} = k_L \frac{{v_G}^2}{L_1} \sin\eta
\end{equation}
\noindent where $\eta$ is the error angle between the aircraft course $\chi$ and the $L_1$ bearing $\chi_L$.
$v_G = \|\mathbf{v}_G\|$, where $\mathbf{v}_G$ is the ground speed vector in inertial frame (North-East 2D plane).
One may set a desired period $P_L$ and damping $\zeta_L$, from which the $L_1$ gain $k_L$ and the $L_1$ ratio $q_L$ may be calculated:
\begin{equation}
\begin{array}{l}
q_L = P_L \zeta_L/\pi \\
k_L = 4{\zeta_L}^2 \\
\end{array}
\end{equation}
\noindent The $L_1$ length may then be adaptively calculated according to the current ground speed:
\begin{equation}
L_1 = q_L v_G
\end{equation}
Using the $L_1$ length,the law of cosines may be used to intermediately determine the angle $\gamma$ between the vector from the aircraft position $\mathbf{p}$ to the center of the circle $\mathbf{c}$ and the $L_1$ vector: 
\begin{equation}
\gamma = \cos^{-1}\left(\operatorname{constrain}\left(\frac{L_1^2 + d^2 - R^2}{ 2 L_1 d},-1,1\right)\right)
\end{equation}
\noindent where distance $d=\|\mathbf{d}\|$, $\mathbf{d}=\mathbf{c}-\mathbf{p}$, is from the aircraft to the loiter center.
$\gamma$ is then used with the loiter direction $s_\text{loit}$ (clockwise: $s_\text{loit}=+1$; counter-clockwise: $s_\text{loit}=-1$) to calculate the $L_1$ bearing:
\begin{equation}
\chi_L = \operatorname{wrap\_pi}\left(\chi_d - s_\text{loit} \gamma\right)
\end{equation} 
\noindent where $\chi_d = \operatorname{atan2}\left(d_e, d_n\right)$ and $\operatorname{atan2}$ is the four-quadrant inverse tangent.
The error angle $\eta$ is then:
\begin{equation}
\eta=\operatorname{constrain}\left(\operatorname{wrap\_pi}\left(\chi_\text{L} - \chi\right),-\frac{pi}{2},\frac{pi}{2}\right)
\end{equation}
\noindent where $\chi = \operatorname{atan2}\left(v_{G_e},v_{G_n}\right)$ is the aircraft course angle and $\operatorname{wrap\_pi}$ wraps its input argument to $\left[-\pi,+\pi\right]$ \SI{}{\radian}.
Now the original acceleration command in~\eqref{eq:aref_orig} can be reformulated as:
\begin{equation}\label{eq:aref_new}
a_\text{ref} = k_L \frac{v_G}{q_L} \sin\eta
\end{equation}
Typical fixed-wing implementations of the algorithm convert the acceleration reference to a roll angle reference $\phi_\text{ref}$, via a coordinated turn assumption, and subsequently saturate the reference:
\begin{equation}
\phi_\text{ref} = \operatorname{constrain}\left(\tan^{-1}\left(\frac{a_\text{ref}}{g}\right),-\phi_\text{lim},\phi_\text{lim}\right)
\end{equation}
\noindent where $g$ is the acceleration of gravity.
Finally -- a detail on implementation -- prevent singularities when aircraft is in center of loiter: e.g. \textbf{if} $d<\epsilon$, \textbf{then} $\mathbf{d}=\left(\epsilon, 0\right)^T$. Any direction may be chosen, or alternatively, one could command zero the acceleration command or hold the previous acceleration command until outside some small radius, $\epsilon$, of the circle center where the standard guidance resumes.
\section{Handling small loiter radii}
Circle following convergence requires that $L_1\leq R$~\cite{L1original}.
Using this relationship, a straight forward adaptation to the $L_1$ ratio may be used whenever this condition is violated (effectively decreasing the operator-defined $L_1$ period).
I.e. the ratio and distance may be recalculated as $q_L = R / v_G$ and $L_1 = R$, respectively.
However, decreasing the period will inherently make the guidance more aggressive, something that may not be desired the when the aircraft is far from the loiter perimeter and turning to approach.
In order to maintain the operator defined gains when possible and adapt only when necessary for convergence to the path, a linear ramp may be applied as the aircraft intercepts the loiter circle using the following logic:
\begin{algorithm}
Calculate nominal operator defined $L_1$ ratio and corresponding distance:\\
$\quad q_L=P_L \zeta_L/\pi$\\
$\quad L_1=q_L v_G$\\
\vspace{0.1cm}
Check loiter convergence criteria and adjust accordingly:\\
\If{$L_1 > R \cap |e_t| \leq L_1$}{
	$L_1 = \operatorname{max}\left(|e_t|,R\right)$\\
	$q_L = L_1 / v_G$\\
}
\noindent where cross track error $e_t=d-R$.
\label{alg:adapt_L1ratio}
\end{algorithm}
Potentially recalculating the $L_1$ ratio requires the following additional check on ground speed: $v_G=\operatorname{max}\left(v_G,v_{G_\text{min}}\right)$ (avoid singularities).
\section{Handling high winds}
\subsection{Bearing feasibility awareness}
\begin{figure}
  \centering
  \begin{subfigure}[b]{0.32\textwidth}
      \includegraphics[width=\textwidth,trim={0 -0.4cm 0 0}]{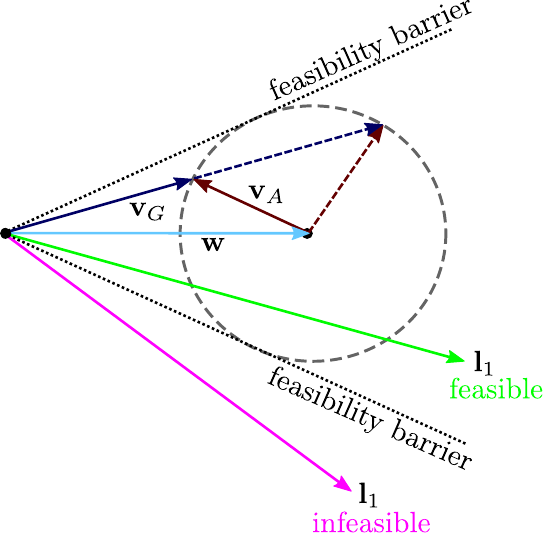}
      \caption{Feasibility ``cone" (wind speed greater than airspeed).}
      \label{fig:feas_cone}
  \end{subfigure}
  \hfill
  \begin{subfigure}[b]{0.64\textwidth}
      \includegraphics[width=\textwidth]{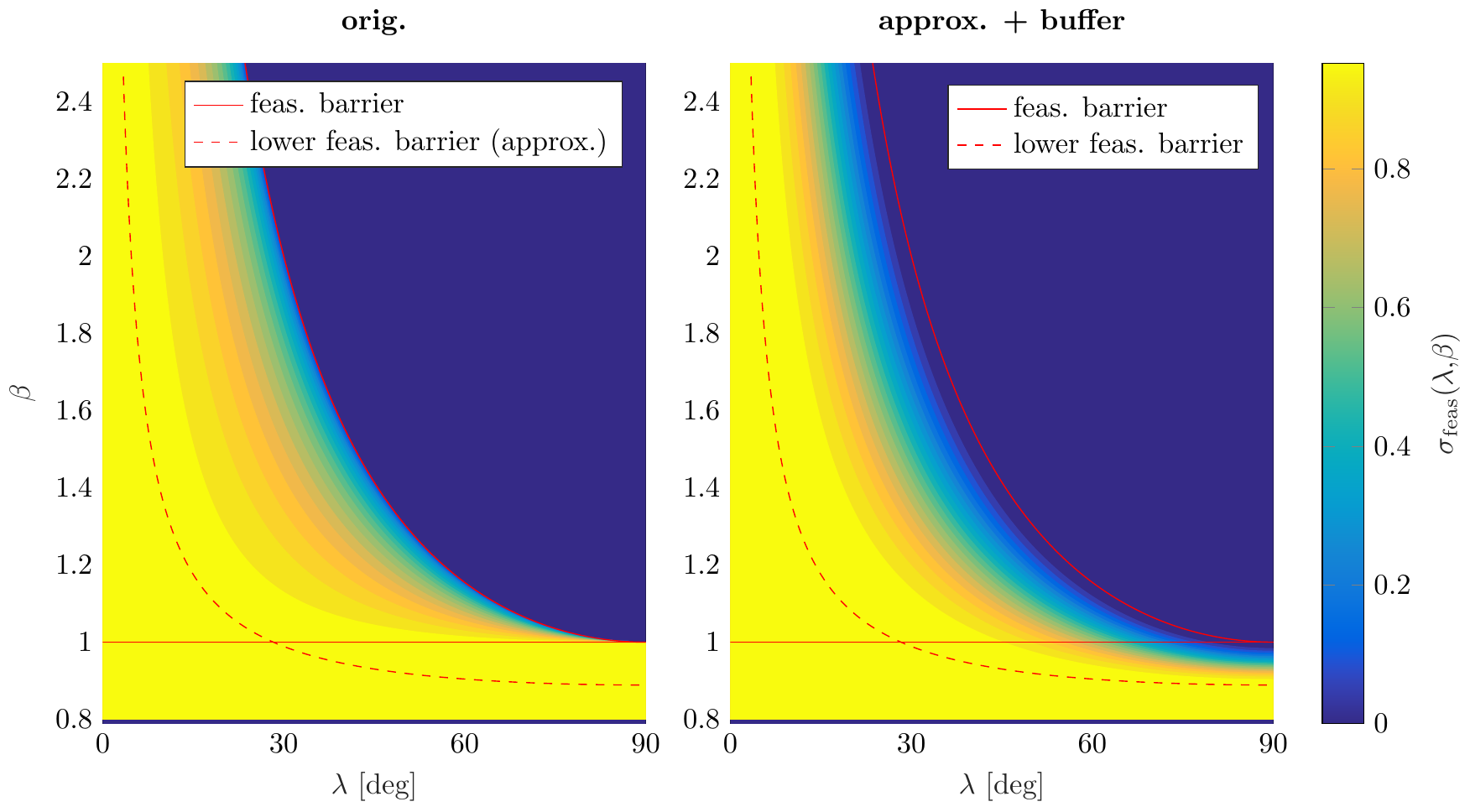}
      \caption{Feasibility function: original formulation from~\cite{Furieri_ACC2017} (left), new approximation with extended buffer zone (right).}
      \label{fig:feas_func}
  \end{subfigure}
  \caption{Bearing feasibility.}\label{fig:feasibility}
\end{figure}

In the case that the wind speed  exceeds the UAV's airspeed, feasibility of flying a given $L_1$ bearing depends on the wind direction.
As described in~\cite{Furieri_ACC2017}, an exact binary boundary on the \emph{bearing feasibility} can be formulated as:
\begin{equation}\label{eq:feas_binary}
\begin{array}{ll}
\beta\sin\lambda\geq 1 \cap |\lambda|\geq\frac{\pi}{2} & \text{(infeasible)} \\
\text{else} & \text{(feasible)} \\
\end{array}
\end{equation}
\noindent where the wind ratio $\beta=w/v_A$ is the fraction of wind speed $w=\|\mathbf{w}\|$ over airspeed $v_A = \|\mathbf{v}_A\| = \|\mathbf{v}_G-\mathbf{w}\|$ and $\lambda$ is the angle between the wind $\mathbf{w}$ and look-ahead $\mathbf{l}_1$ vectors, $\in\left[-\pi,\pi\right]$.
\begin{equation}
\lambda = \operatorname{atan2}\left(\mathbf{w}\times\mathbf{l}_1,\mathbf{w}\cdot\mathbf{l}_1\right)
\end{equation}
\noindent where $\mathbf{l}_1=L_1\left(\cos\chi_L, \sin\chi_L\right)^T$.
The relationship in~\eqref{eq:feas_binary} physically describes a ``feasibility cone", asymptotically decreasing to zero angular opening as the wind ratio increases above unity, see Fig.~\ref{fig:feas_cone}.
When the $\mathbf{l}_1$ vector lies within this cone the bearing is feasible, and contrarily when outside, infeasible.
As outlined in~\cite{Furieri_ACC2017}, two separate tracking objectives can then be intuited: 1) an \emph{ideal} tracking objective, where we are able to track the prescribed bearing and 2) a \emph{safety} objective, where we instead tend towards reducing run-away by turning against the wind and simultaneously leveling the aircraft as $t\rightarrow \infty$, where $t$ is time.
As binary steps in tracking objectives will cause oscillations in guidance commands when the vehicle remains on or near the feasibility boundary (common when the wind is approaching the airspeed and small gusts, wind shear, or turbulence is present), it is further desirable to transition continuously through these two states.
In~\cite{Furieri_ACC2017}, the following transitioning function (equivalently, continuous feasibility function) was proposed:
\begin{equation}\label{eq:feas_old}
\sigma_\text{feas} = \frac{\sqrt{1-\left(\beta\sin\lambda\right)^2}}{cos\lambda}
\end{equation}
\noindent where $\beta<1$ (wind speed is less than airspeed) and $\lambda=0$ \SI{}{\radian} (bearing is aligned with wind direction) result in \emph{feasible} output $\sigma_\text{feas}=1$, states beyond the feasibility boundary result in \emph{infeasible} output $\sigma_\text{feas}=0$, and a continuous function $\sigma_\text{feas}\in\left[0,1\right]$ in between, see Fig.~\ref{fig:feas_func} (left).
Some practical issues exist, however, with the function as defined in~\eqref{eq:feas_old}; namely, the transition is continuous but not smooth at the feasibility boundary, which can lead to jagged reference commands, and further, numerical stability issues exist as $\lambda\rightarrow\frac{\pi}{2} \cap \beta\rightarrow 1$.
To address these issues, a small buffer zone below the $\beta=1$ line may be designed, considering some buffer airspeed $v_{A_\text{buf}}$.
\begin{equation}
\beta_\text{buf} =  v_{A_\text{buf}} /  v_A
\end{equation}
\noindent The buffer's magnitude may be set to a reasonable guess at the wind estimate or airspeed uncertainty, or, as outlined in more detail in the following section, set depending on the airspeed reference tracking dynamics (e.g. a conservative buffer in which the airspeed reference may be properly tracked by the end of the transition).
Additionally, an approximation of the feasibility function in~\eqref{eq:feas_old} can be made incorporating the buffer zone, as well as maintaining continuity and smoothness in the transition, see eq.~\eqref{eq:feas_new} and Fig.~\ref{fig:feas_func} (right).
\begin{equation}\label{eq:feas_new}
\sigma_\text{feas}= \begin{cases}
0 & \beta > \beta_+ \\
\cos\left(\frac{\pi}{2}\operatorname{constrain}\left(\frac{\beta-\beta_-}{\beta_+-\beta_-},0,1\right)\right)^2 & \beta > \beta_- \\
1 & \text{else}
\end{cases}
\end{equation}
where the upper limit of the transitioning region $\beta_+$ is approximated as a piecewise function with a linear finite cut-off to avoid singularities, the cut-off angle $\lambda_\text{co}$ chosen such that the regular operational envelope is not affected:
\begin{equation}
\beta_+ = \begin{cases}
\beta_{+_\text{co}} + m_\text{co}\left(\lambda_\text{co}-\lambda_\text{ctsr}\right) & \lambda_\text{ctsr} < \lambda_\text{co} \\
1/\sin\lambda_\text{ctsr} & \text{else}
\end{cases}
\end{equation}
\noindent with $\beta_{+_\text{co}} = 1/\sin\lambda_\text{co}$, $m_\text{co} = \cos\lambda_\text{co}/\sin\lambda_\text{co}^2$, and $\lambda_\text{ctsr}=\operatorname{constrain}\left(|\lambda|,0,\frac{\pi}{2}\right)$. The lower limit of the transitioning region $\beta_-$ is similarly made piecewise to correspond with $\beta_+$:
\begin{equation}
\beta_- = \begin{cases}
\beta_{-_\text{co}} + \beta_\text{buf} m_\text{co}\left(\lambda_\text{co}-\lambda_\text{ctsr}\right) & \lambda_\text{ctsr} < \lambda_\text{co} \\
\left(1/\sin\lambda_\text{ctsr}-2\right)\beta_\text{buf} + 1 & \text{else} 
\end{cases}
\end{equation}
\noindent where $\beta_{-_\text{co}} = \left(1/\sin\lambda_\text{co}-2\right)\beta_\text{buf} + 1$.
With a new feasibility function defined, it's application to the $L_1$ algorithm can be elaborated; specifically, it is desired that 1) when the bearing is feasible, $L_1$ operates as usual, 2) when the bearing is infeasible, $L_1$ transitions to the \emph{safety} objective, and 3) in between these states, the reference commands maintain continuity and avoid oscillations.
The safety objective can be achieved by replacing the ground speed vector $\mathbf{v}_G$ with the airspeed vector $\mathbf{v}_A$ in eq.~\eqref{eq:aref_new}.
Continuity may the be obtained utilizing the feasibility function $\sigma_\text{feas}$ in~\eqref{eq:feas_new} as follows:
\begin{equation}
\mathbf{v}_\text{nav} = \sigma_\text{feas}\mathbf{v}_G + \left(1-\sigma_\text{feas}\right)\mathbf{v}_A
\end{equation}
\noindent subsequently calculating the error angle
\begin{equation}
\eta=\operatorname{constrain}\left(\operatorname{wrap\_pi}\left(\chi_\text{nav} - \chi\right),-\frac{pi}{2},\frac{pi}{2}\right)
\end{equation}
\noindent where $\chi_\text{nav} = \operatorname{atan2}\left(v_{\text{nav}_e},v_{\text{nav}_n}\right)$, and computing the final acceleration reference
\begin{equation}
a_\text{ref} = k_L \frac{v_\text{nav}}{q_L} \sin\eta
\end{equation}
The resulting behavior of the aircraft will \emph{mitigate} run-away scenarios; i.e., in over-wind scenarios, minimize the run-away as much as possible at a single commanded airspeed.
However, maximum airspeed allowing, it is also possible to \emph{prevent} run-away from the track completely via \emph{airspeed reference compensation}.
The next section details a simple approach towards this end.
\subsection{Airspeed reference compensation}
Though nominal airspeed references are often desired for energy efficiency, in critical conditions, e.g. very high winds, airspeed reference increases may be allowed either while short-term gusts or wind shear persists, or until an emergency landing may be executed.
The most straight forward approach would be to match whatever wind speed overshoot (w.r.t. the airspeed) with airspeed reference commands.
However, as previously detailed, the wind magnitude does not alone dictate the bearing feasibility; more so, the relation between wind direction and desired bearing.
Utilizing the bearing feasibility function (eq.~\eqref{eq:feas_new}) defined in the prior sections, the vehicle may more appropriately command airspeed increments.
With a nominal and maximum airspeed reference defined, $v_{A_\text{nom}}$ and $v_{A_\text{max}}$, respectively, a wind speed and bearing feasibility dependent airspeed reference increment may be calculated.
\begin{equation}
\Delta v_{A_w} = \operatorname{constrain}\left(w-v_{A_\text{nom}}, 0, \Delta v_{A_\text{max}}\right)\left(1 - \sigma_\text{feas}\right)
\end{equation}
\noindent where $\Delta v_{A_\text{max}} = v_{A_\text{max}}-v_{A_\text{nom}}$ is the maximum allowed airspeed reference increment.
The airspeed reference then computed as
\begin{equation}
v_{A_\text{ref}} = v_{A_\text{nom}} + \Delta v_{A_w}
\end{equation}
\noindent Note the buffer zone is essential here, as the resultant equilibrium point of this algorithm approaches $\beta=1$ and $\lambda=\frac{\pi}{2}$, i.e. zero ground speed, and facing into the wind.
\section{Example simulations}
The following simulations demonstrate the outlined $L_1$ extensions within this brief. All simulations were executed in MATLAB with a simplified 2D model of a small UAV, with first order airspeed and roll angle dynamics, as follows:
\begin{equation}
\left(\begin{matrix}
\dot{n} \\ \dot{e} \\ \dot{v_A} \\ \dot{\xi} \\ \dot{\phi}
\end{matrix}\right)=\left(\begin{matrix}
v_A\cos\xi + w_n \\ v_A\sin\xi + w_e \\ \left(v_{A_\text{ref}}-v_A\right)/\tau_v \\ g\tan\phi / v_A \\ \left(\phi_\text{ref}-\phi\right) / \tau_\phi
\end{matrix}\right)
\end{equation}
\noindent where heading $\xi=\operatorname{atan2}\left(v_{A_e},v_{A_n}\right)$ and the time constants $\tau_v=1$\SI{}{\second} and $\tau_\phi=0.5$\SI{}{\second} are representative of a small, slow speed radio-controlled UAV, running standard low-level attitude stabilization (PID) and e.g. TECS (Total Energy Control System) for airspeed/altitude control.
Wind dynamics are detailed in each respective simulation.
Guidance parameters for all simulations are held constant, values listed in Table~\ref{tab:guidance_params}.
\begin{table}
\centering
\caption{Guidance parameters used in simulations.}
\label{tab:guidance_params}
\begin{tabular}{rl|rl}
Param & Value & Param & Value\\
$P_L$ & \SI{25}{\meter} & $v_{A_\text{nom}}$ & \SI{9}{\meter\per\second} \\
$\zeta_L$ & \SI{0.707}{} & $v_{A_\text{max}}$ & \SI{12}{\meter\per\second} \\
$\phi_\text{lim}$ & \SI{35}{\degree} & $v_{A_\text{buf}}$ & \SI{1}{\meter\per\second} \\
\end{tabular}
\end{table}
\subsection{Handling small loiter radii}
\begin{figure}[h]
  \centering
  \includegraphics[width=1\textwidth,trim={0.5cm 0 0.5cm 0},clip]{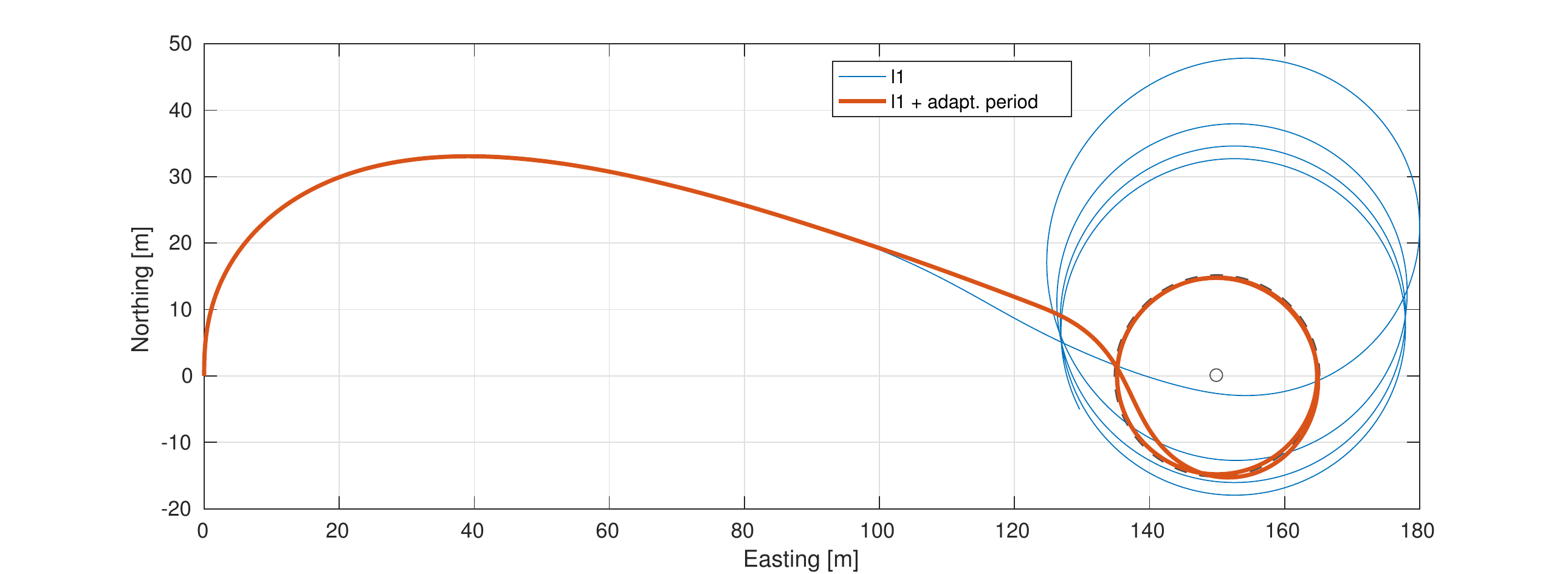}
  \caption{Following a loiter circle with a small (w.r.t. the flight speed) radius of \SI{15}{\meter}. Adapting the $L_1$ ratio accordingly allows path convergence.}\label{fig:small_radii-pos}
\end{figure}

\begin{figure}[h]
  \centering
  \includegraphics[width=1\textwidth,trim={0.5cm 0 0.5cm 0},clip]{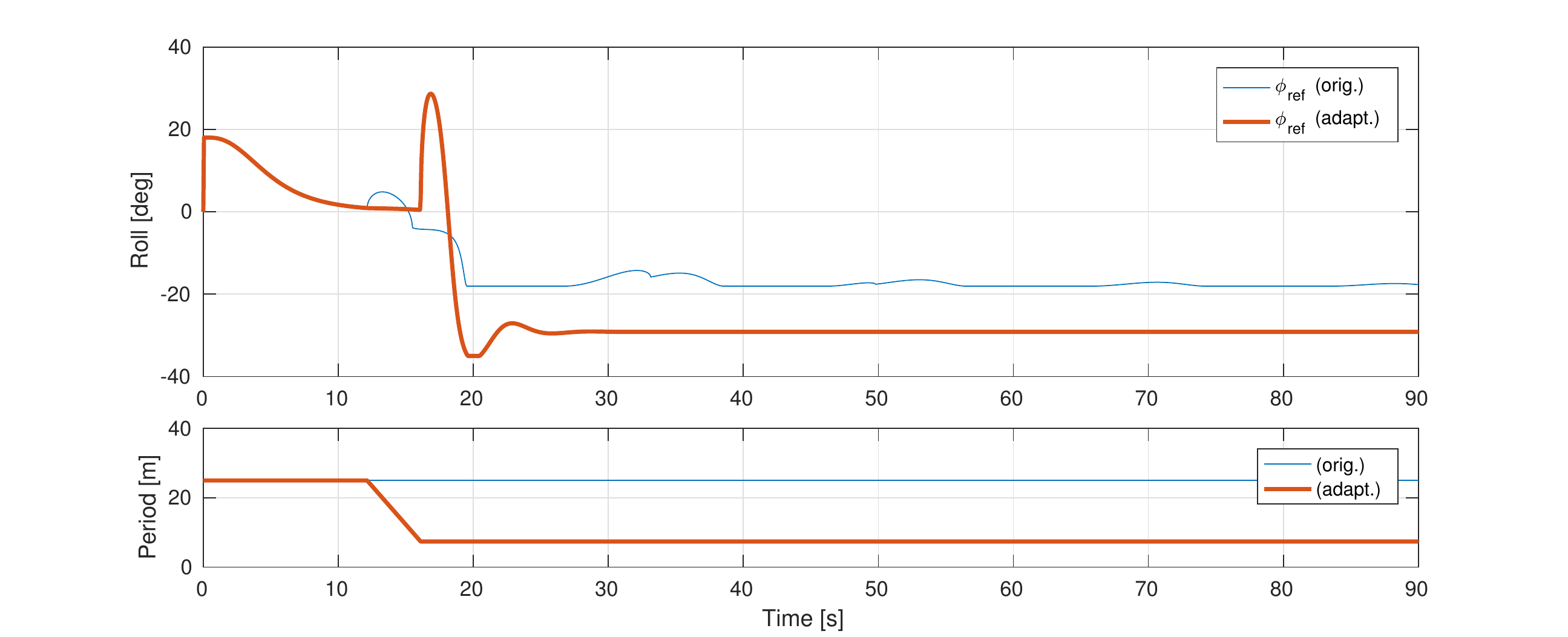}
  \caption{Roll references and the $L_1$ period while following a small loiter.}\label{fig:small_radii-states}
\end{figure}

\begin{figure}[h]
  \centering
  \includegraphics[width=1\textwidth,trim={0.5cm 0 0.5cm 0},clip]{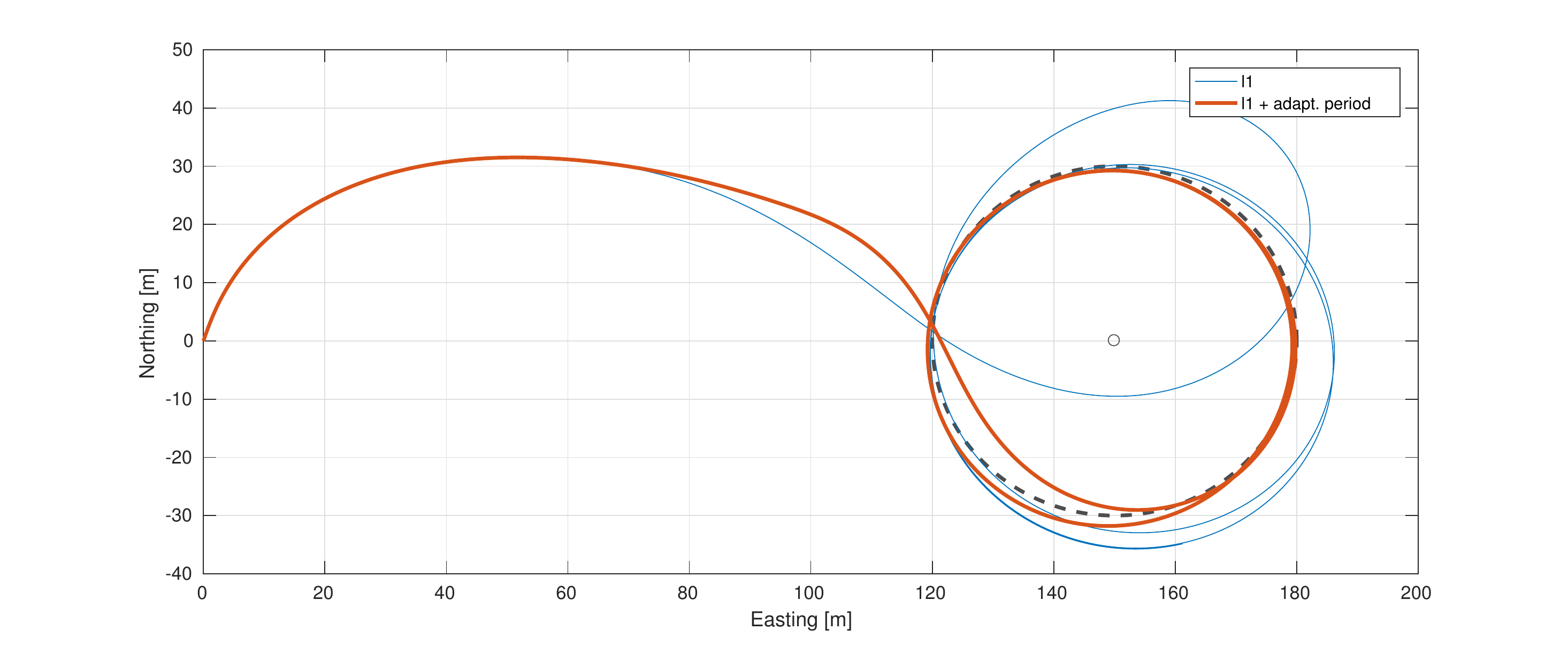}
  \caption{Loitering in in \SI{3}{\meter\per\second} eastward wind. The $L_1$ ratio adapts to keep closer to the track.}\label{fig:small_radii-wind-pos}
\end{figure}

\begin{figure}[h]
  \centering
  \includegraphics[width=1\textwidth,trim={0.5cm 0 0.5cm 0},clip]{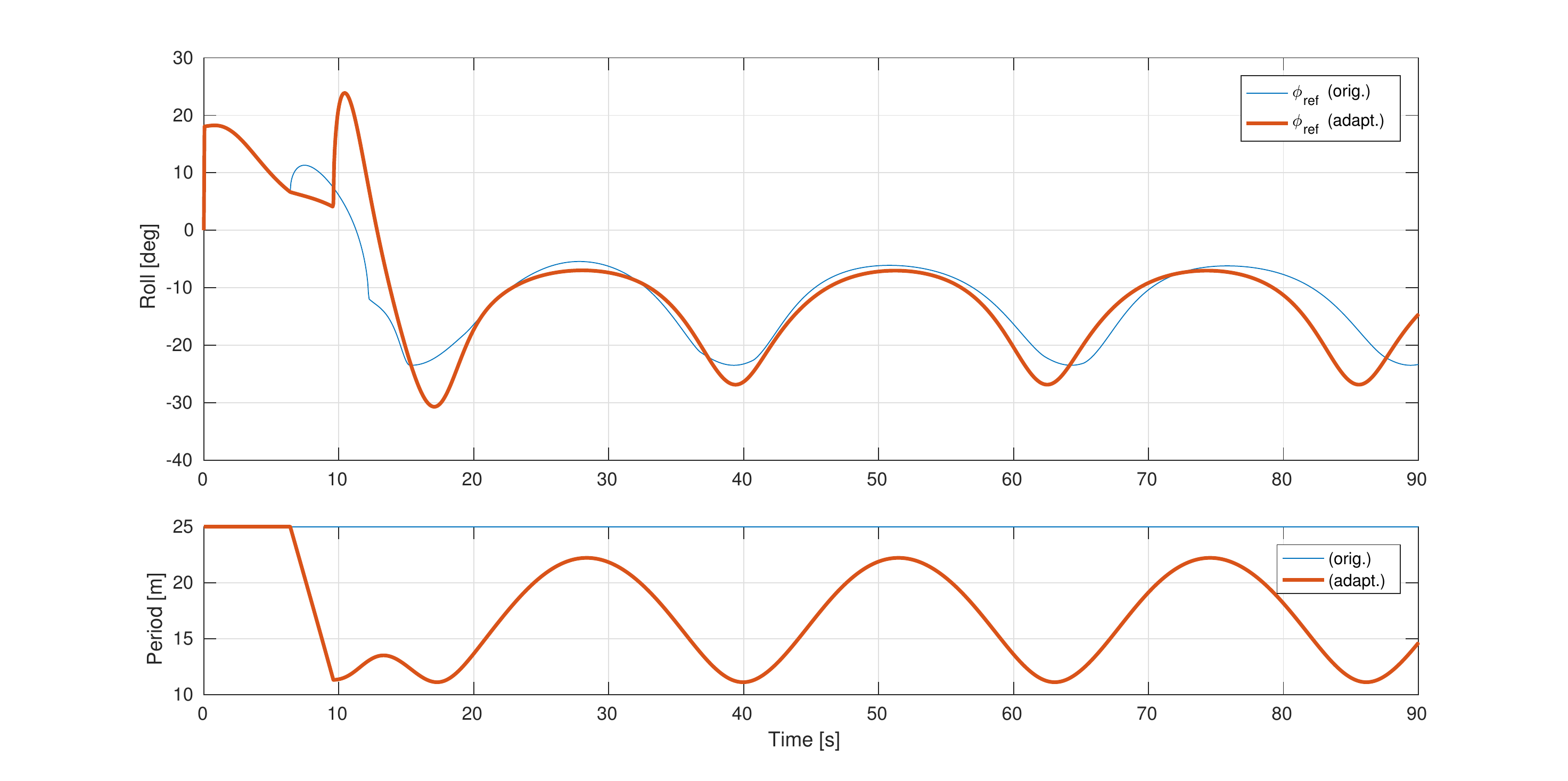}
  \caption{Roll references and the effective $L_1$ period while following a loiter in \SI{3}{\meter\per\second} eastward wind.}\label{fig:small_radii-wind-states}
\end{figure}

Figures~\ref{fig:small_radii-pos} and~\ref{fig:small_radii-states} show the effect of implementing the adaptive $L_1$ ratio vs. the original formulation.
A small (relative to the flight speed) $R=$\SI{15}{\meter} radius loiter is followed in no wind by the adaptive formulation, by effectively reducing the $L_1$ period, while the non-adaptive formulation does not converge to the path.
Figures~\ref{fig:small_radii-wind-pos} and~\ref{fig:small_radii-wind-states} show the effect of the adaptive $L_1$ ratio further in moderate (\SI{3}{\meter\per\second}) eastward wind.
As the ratio is proportional to ground speed, the effective period will rise and fall corresponding to the orientation to the wind, maintaining closer loiter tracking than in the non-adapted case.
\subsection{Handling high winds}
\begin{figure}[h]
  \centering
  \includegraphics[width=1\textwidth,trim={0.5cm 0 0.5cm 0},clip]{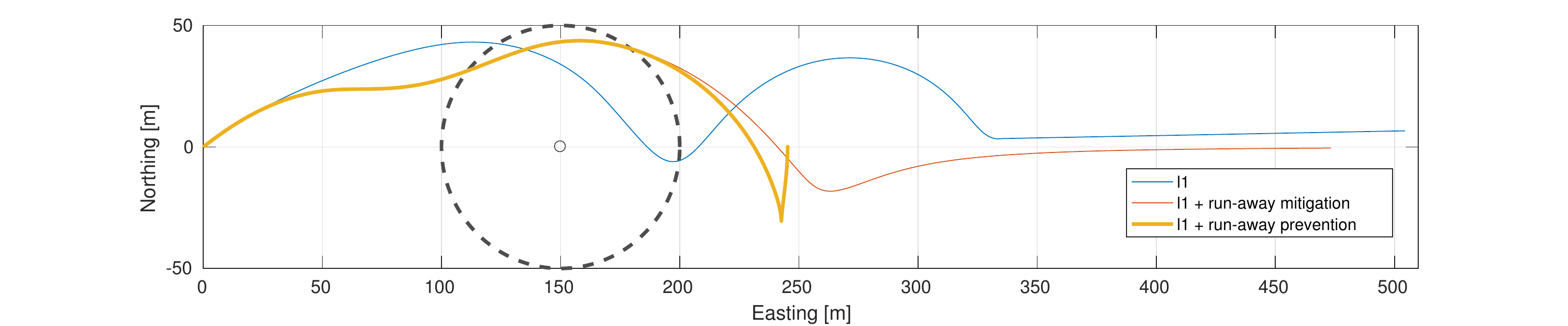}
  \caption{Run-away \emph{mitigation} and \emph{prevention} in a constant over-wind.}\label{fig:const_wind-pos}
\end{figure}

\begin{figure}[h]
  \centering
  \includegraphics[width=1\textwidth,trim={0.5cm 1.5cm 0.5cm 0},clip]{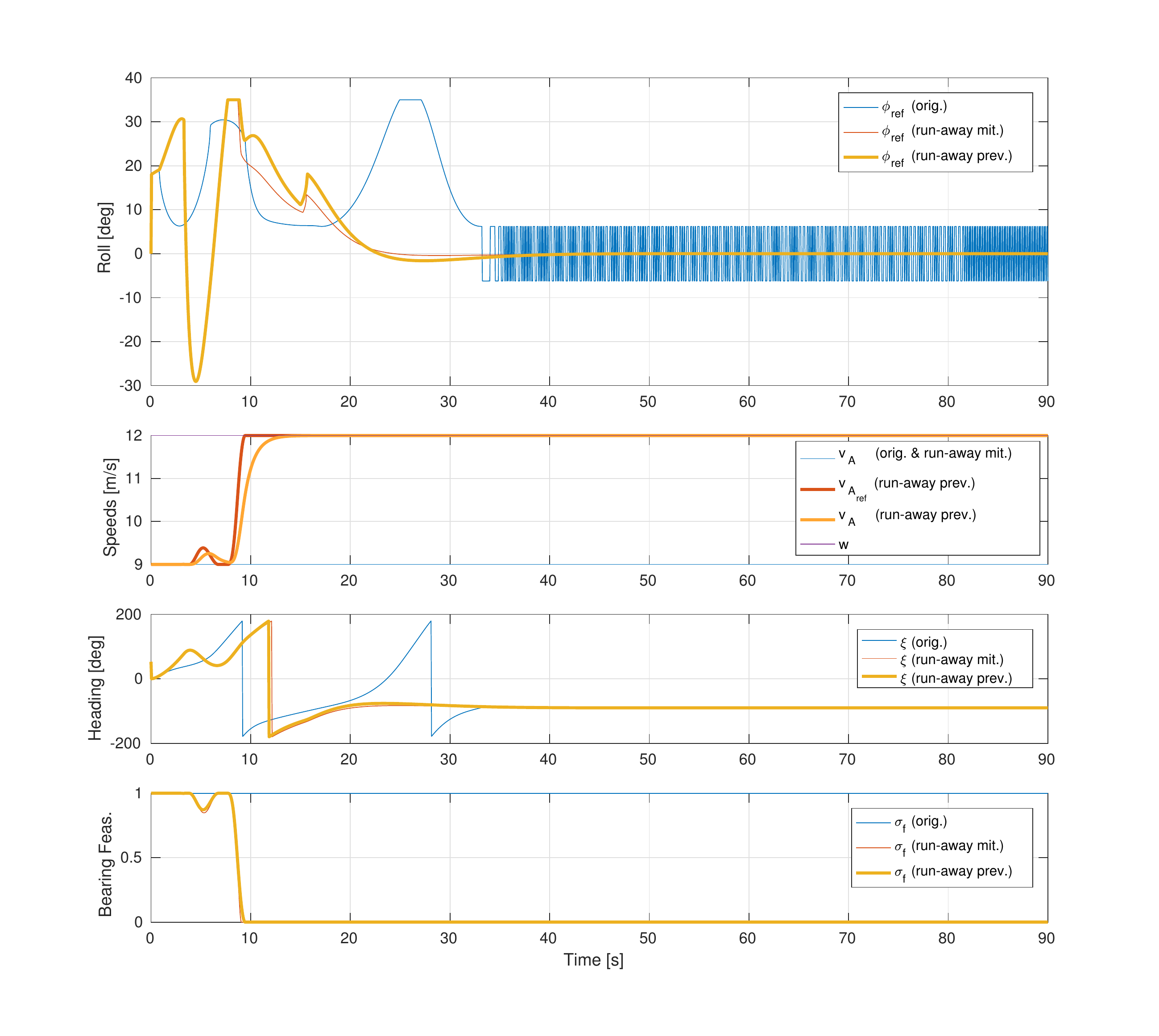}
  \caption{Aircraft states and controls for loitering in a constant over-wind.}\label{fig:const_wind-states}
\end{figure}

\begin{figure}[h]
  \centering
  \includegraphics[width=1\textwidth,trim={0.5cm 0cm 0.5cm 0},clip]{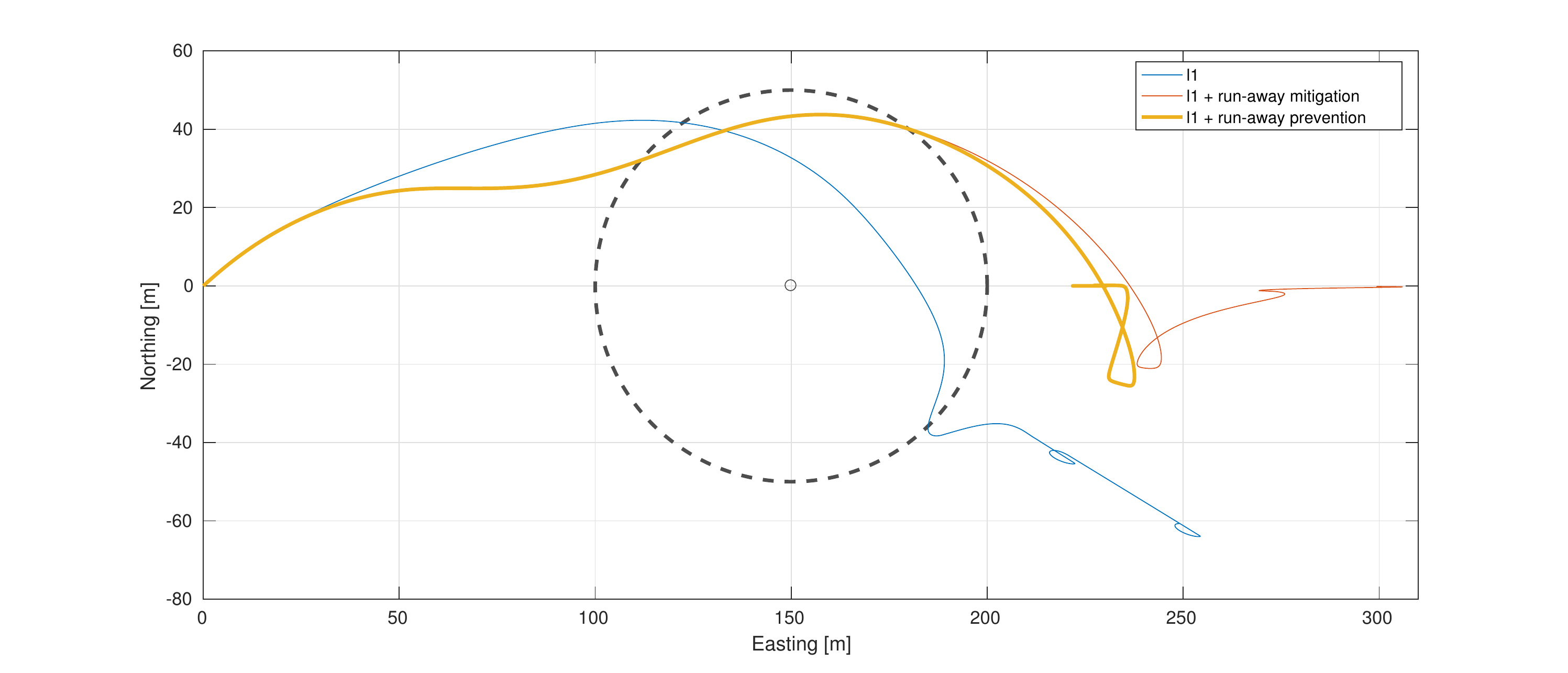}
  \caption{Run-away \emph{mitigation} and \emph{prevention} in a constant over-wind with sinusoidal gusting.}\label{fig:sin_wind-pos}
\end{figure}

\begin{figure}[h]
  \centering
  \includegraphics[width=1\textwidth,trim={0.5cm 1cm 0.5cm 0},clip]{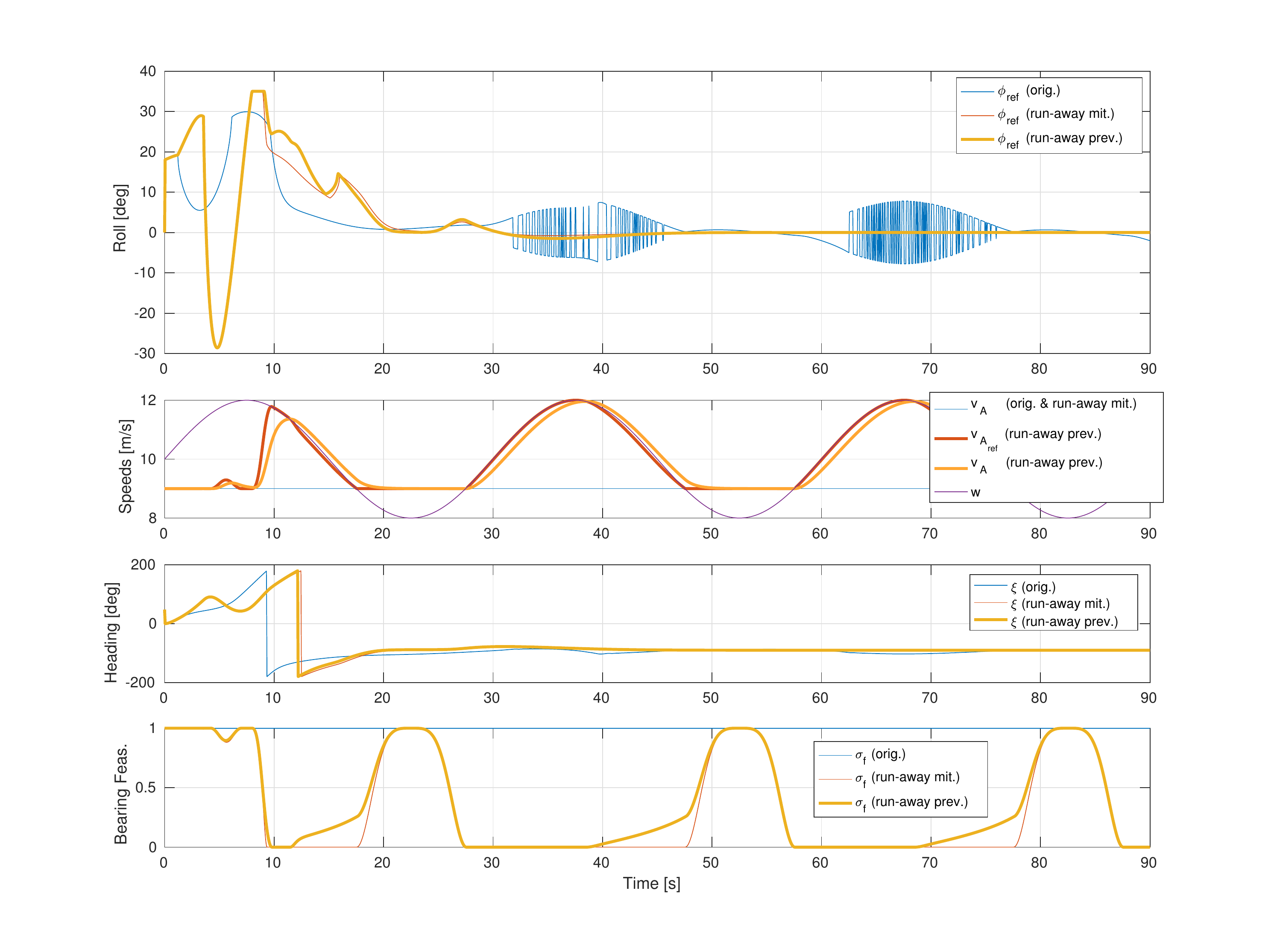}
  \caption{Aircraft states and controls for loitering in a constant over-wind with sinusoidal gusting.}\label{fig:sin_wind-states}
\end{figure}

In figures~\ref{fig:const_wind-pos} and~\ref{fig:const_wind-states}, the bearing feasibility function is introduced for the purposes of both run-away \emph{mitigation} and \emph{prevention} with a constant eastward wind of \SI{12}{\meter\per\second} (\SI{3}{\meter\per\second} over the commanded airspeed).
In the original case, the $L_1$ logic is not able to account for the ``backwards" flight motion with respect to the ground, as it only considers the ground velocity vector, leading to run-away with large roll reference bang-bang oscillations, due to the unstable operating point it converges to.
Note in practice, aside from the obviously undesirable command oscillations, these bang-bang controls combined with any other small perturbations (gusts or couple motions from the longitudinal/direction axes) will often lead to ``turn-around" trajectories, which can be disconcerting to operators viewing the flight from the ground.
With run-away \emph{mitigation} active, the bearing feasibility function schedules continuous control action towards tracking the loiter when feasible, and turn against the wind (safety) when infeasible, resulting in the slowest run-away velocity configuration with a stable reference command.
Once run-away \emph{prevention} is active, the airspeed reference is allowed to incrementally increase with the feasibility parameter to match the wind speed in the infeasible case, and remain at nominal when tracking is feasible.
This results in a zero ground speed terminal configuration, facing against the wind.
In figures~\ref{fig:sin_wind-pos} and~\ref{fig:sin_wind-states}, a \SI{2}{\meter\per\second} amplitude, \SI{30}{\second} period sinusoidal eastward wind gust is introduced varying about a constant eastward wind of \SI{10}{\meter\per\second}.
Similar bang-bang reference roll oscillations are seen in the infeasible bearing cases for the original formulation, while both the run-away mitigation and prevention results similarly maintain continuous and safe control commands.
The airspeed reference can be seen to follow the wind speed as it increases above the nominally command reference threshold and the bearing becomes infeasible.
%


\bibliography{lib}
\bibliographystyle{aiaa}

\end{document}